\documentclass{article}
\usepackage{spconf,amsmath,graphicx}

\usepackage{url}
\usepackage{graphicx}
\usepackage{subfigure}

\title{Robust Full-FoV Depth Estimation in Tele-wide Camera System}

\name{Kai Guo, Seongwook Song, Soonkeun Chang, Tae-ui Kim, Seungmin Han and Irina Kim}
   \vspace{-0.3cm}

\address{Samsung Electronics, Hwaseong-si, Korea}

   \vspace{-0.4cm}
\begin{document}

\maketitle

\begin{abstract}
   \vspace{-0.15cm}
Tele-wide camera system with different Field of View (FoV) lenses becomes very popular in recent mobile devices. Usually it is difficult to obtain full-FoV depth based on traditional stereo-matching methods. Pure Deep Neural Network (DNN) based depth estimation methods can obtain full-FoV depth, but have low robustness for scenarios which are not covered by training dataset. In this paper, to address the above problems we propose a hierarchical hourglass network for robust full-FoV depth estimation in tele-wide camera system, which combines the robustness of traditional stereo-matching methods with the accuracy of DNN. More specifically, the proposed network comprises three major modules: single image depth prediction module infers initial depth from input color image, depth propagation module propagates traditional stereo-matching tele-FoV depth to surrounding regions, and depth combination module fuses the initial depth with the propagated depth to generate final output. Each of these modules employs an hourglass model, which is a kind of encoder-decoder structure with skip connections. Experimental results compared with state-of-the-art depth estimation methods demonstrate that our method not only produces robust and better subjective depth quality on wild test images, but also obtains better quantitative results on standard datasets.
\end{abstract}


   \vspace{-0.3cm}
\section{Introduction}
\label{sec:intro}

   \vspace{-0.2cm}

Currently tele-wide camera system with different Field of View (FoV) lenses is very popular in mobile devices, wherein each lens has different FoV, e.g. wide angle lens,  tele zoom lens etc. How to obtain robust full-FoV depth for this kind of camera system becomes a challenging problem.

Usually it is difficult to obtain full-FoV depth by traditional stereo-matching methods \cite{Scharstein02:IJCV, Zhan16:CSVT}. This kind of method can be formulated as a three-step pipeline including matching cost calculation \cite{Hernandez16:PCS}, cost aggregation/optimization \cite{Hirschmuller08:PAMI, Zhang07:PAMI}, and disparity refinement \cite{Barron16:ECCV, Min14:TIP}.

Recently Deep Neural Network (DNN) has been successfully applied to the single image depth prediction \cite{Eigen14:NIPS, Xie16:ECCV, Laina16:3DV, Godard17:CVPR, Fu18:CVPR, Li18:CVPR} and tele-wide stereo depth estimation \cite{El-Khamy19:CVPRW}. Eigen et al. \cite{Eigen14:NIPS} combined a coarse global prediction network based on the entire image with a refinement network, and proposed a scale-invariant L2-norm loss function which is widely utilized in depth prediction. Laina et al. \cite{Laina16:3DV} developed an up-projection decoder and achieved higher accuracy. Godard et al. \cite{Godard17:CVPR} proposed an unsupervised method to enforce disparity consistency between the left and right images, achieving similar depth quality compared with supervised methods. Li et al. \cite{Li18:CVPR} presented a large depth dataset called MegaDepth, which is generated from collected multi-view internet photos, the hourglass model trained on this dataset shows good results on wild test images.  EI-Khamy et al. \cite{El-Khamy19:CVPRW} proposed a stereo matching neural network for tele-wide camera system so as to estimate the full-FoV depth.

\begin{figure}[t]
    \begin{minipage}[b]{0.32\linewidth}
      %\centering
      \centerline{\includegraphics[width= 1.2in]{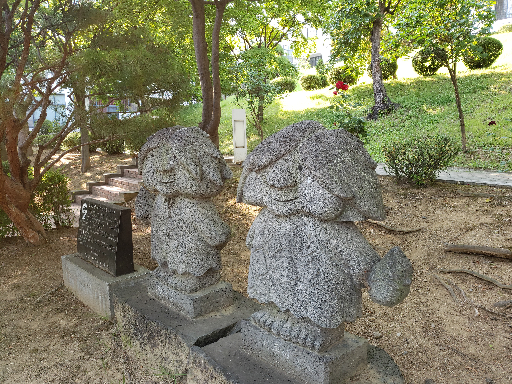}}
      \vspace{0.00cm}
      \centerline{(a)}\medskip
    \end{minipage}
    \hfill
    \begin{minipage}[b]{0.15\linewidth}
     \centerline{\raisebox{0.5\height}{\includegraphics[width= 0.58in]{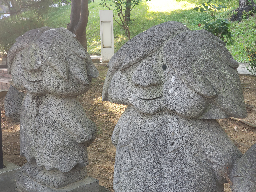}}}
      \vspace{0.00cm}
      \centerline{(b)}\medskip
    \end{minipage}
    \hfill
    \begin{minipage}[b]{0.15\linewidth}
     \centerline{\raisebox{0.5\height}{\includegraphics[width= 0.58in]{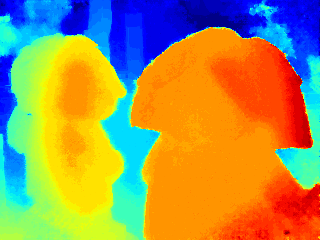}}}
      \vspace{0.00cm}
      \centerline{(c)}\medskip
    \end{minipage}
    \hfill
    \begin{minipage}[b]{0.32\linewidth}
     \centering
      \centerline{\includegraphics[width= 1.2in]{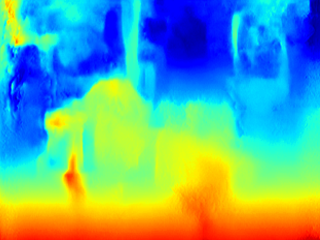}}
      \vspace{0.00cm}
      \centerline{(d)}\medskip
    \end{minipage}

    \begin{minipage}[b]{0.32\linewidth}
      %\centering
      \centerline{\includegraphics[width= 1.2in]{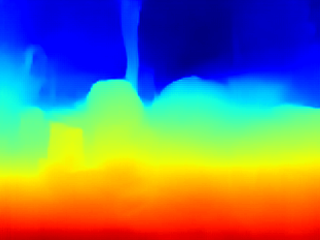}}
      \vspace{0.00cm}
      \centerline{(e)}\medskip
    \end{minipage}
    \hfill
    \begin{minipage}[b]{0.32\linewidth}
      %\centering
      \centerline{\includegraphics[width= 1.2in]{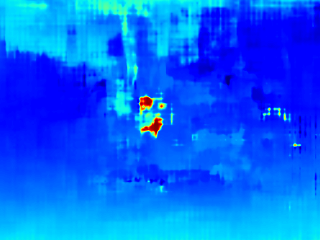}}
      \vspace{0.00cm}
      \centerline{(f)}\medskip
    \end{minipage}
    \hfill
    %\hspace{1.46cm}
    \begin{minipage}[b]{0.32\linewidth}
      %\centering
      \centerline{\includegraphics[width= 1.2in]{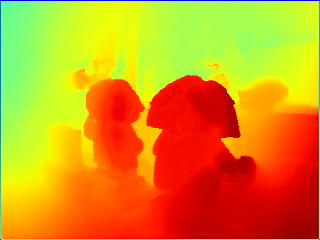}}
      \vspace{0.00cm}
      \centerline{(g)}\medskip
    \end{minipage}

   \vspace{-0.4cm}

    %\centerline{\epsfig{figure=msePS.eps,width=3.0in}}
    \caption {Depth comparisons. (a) Wide image; (b) Tele image; (c) Traditional tele-FoV stereo-matching method: matching cost calculation/optimization \cite{Hernandez16:PCS} + post processing \cite{Min14:TIP}; (d) DNN-based single image depth method in \cite{Godard17:CVPR}; (e) DNN-based single image depth method in \cite{Li18:CVPR}; (f) Pure DNN-based tele-wide stereo matching method in \cite{El-Khamy19:CVPRW}; (g) Our result.} \label{fig:expe_wild_statue}
   \vspace{-0.4cm}
    \end{figure}

%However, recent studies showed the lack of robustness in well-trained deep neural networks in image understanding \cite{Su18:ECCV}. 
% For depth prediction, the state-of-the-art well trained DNN model can not generalize well in wild test scenarios, especially the scenarios which not covered by training dataset

Pure DNN-based methods usually have low robustness for scenarios which are not covered by training dataset \cite{Nguyen18:ECCV, Su18:ECCV} (as shown in Fig. \ref{fig:expe_wild_statue} (d, e and f)). To address the above problems, we propose a hierarchical hourglass model, which incorporates the traditional tele-FoV stereo-matching depth as input, and estimate more robust full-FoV depth on various test scenarios (as shown in Fig. \ref{fig:expe_wild_statue} (g)). More specifically, the proposed network comprises three modules: single image depth prediction module that infers initial depth from input color image, depth propagation module that propagates traditional stereo-matching depth from tele-FoV to surrounding regions, and depth combination module that fuses the initial depth with the propagated depth. All of them employ the hourglass model \cite{Chen16:NIPS}, which has an encoder-decoder structure with skip connections. Experiments demonstrate that our method can not only get robust and better subjective depth quality than state-of-the-art depth estimation on wild test images, but also obtain better objective results on standard datasets.

\begin{figure}
\centerline{\includegraphics[width=\columnwidth]{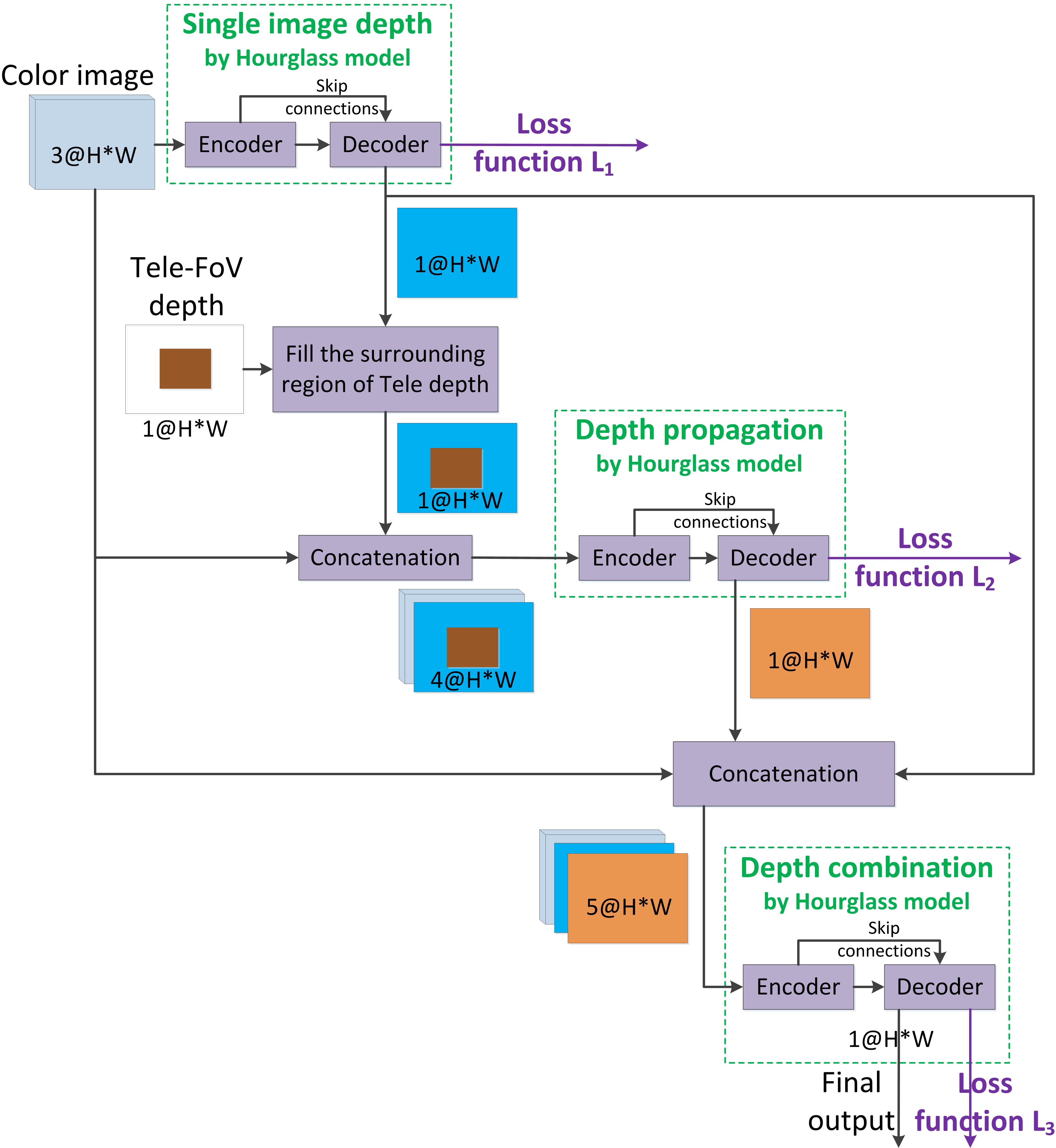}}
   \vspace{-0.3cm}
\caption{Architecture of the proposed Hierarchical Hourglass network. It comprises three major modules: single image depth prediction, depth propagation and depth combination. All of these modules employ the Hourglass model \cite{Chen16:NIPS}. The loss function of network is weighted sum of all these modules loss functions.}
   \vspace{-0.4cm}
\label{fig:flowchart_HieHourglass}
\end{figure}

\begin{figure}[t]
    \begin{minipage}[b]{0.32\linewidth}
      %\centering
      \centerline{\includegraphics[width= 1.2in]{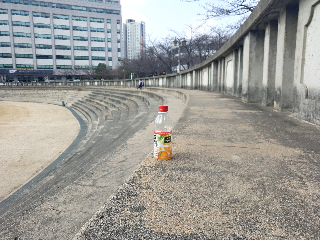}}
      \vspace{0.00cm}
      \centerline{(a)}\medskip
    \end{minipage}
    \hfill
    %\hspace{1.46cm}
    \begin{minipage}[b]{0.32\linewidth}
     \centering
      %\centerline{\includegraphics[width= 0.6in]{wild_stadium_2_teleDepth.png}}
      \centerline{\raisebox{0.43\height}{\includegraphics[width= 0.63in]{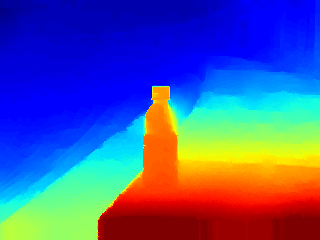}}}
      \vspace{0.00cm}
      \centerline{(b)}\medskip
    \end{minipage}
   \hfill
    %\hspace{-5.8cm}
    \begin{minipage}[b]{0.32\linewidth}
      % \centering
      %\raisebox{-0.8\height}
      \centerline{\includegraphics[width= 1.2in]{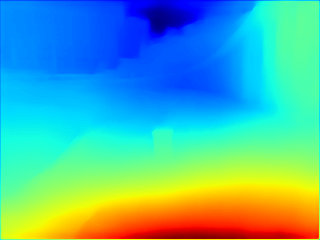}}
      \vspace{0.00cm}
      \centerline{(c)}\medskip
    \end{minipage}

    \begin{minipage}[b]{0.32\linewidth}
      %\centering
      \centerline{\includegraphics[width= 1.2in]{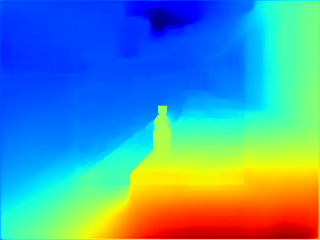}}
      \vspace{0.00cm}
      \centerline{(d)}\medskip
    \end{minipage}
    \hfill
    \begin{minipage}[b]{0.32\linewidth}
      %\centering
      \centerline{\includegraphics[width= 1.2in]{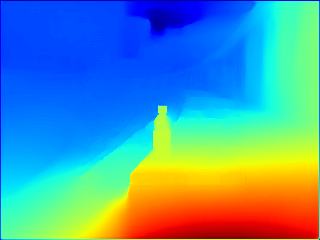}}
      \vspace{0.00cm}
      \centerline{(e)}\medskip
    \end{minipage}
    \hfill
    %\hspace{1.46cm}
    \begin{minipage}[b]{0.32\linewidth}
      %\centering
      \centerline{\includegraphics[width= 1.2in]{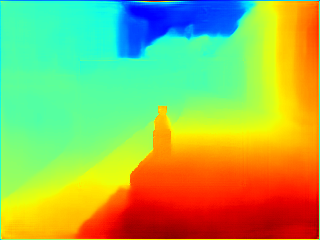}}
      \vspace{0.00cm}
      \centerline{(f)}\medskip
    \end{minipage}

   \vspace{-0.4cm}

    \caption {Results of our network which employs the proposed L1-norm scale-invariant loss function: (c, d and e), 
and result of our network which employs widely-used L2-norm scale-invariant loss function \cite{Eigen14:NIPS}: (f). 
(a) Wide image of input; (b) Tele-FoV stereo depth of input (estimated by matching cost calculation/optimization \cite{Hernandez16:PCS} + post processing \cite{Min14:TIP}); (c) Intermediate result of single image depth module; (d) Intermediate result of depth propagation module; (e) Final depth after depth combination module.} \label{fig:expe_wild_stadium}
   \vspace{-0.4cm}

    \end{figure}

\begin{figure}
\centerline{\includegraphics[width=2.0in]{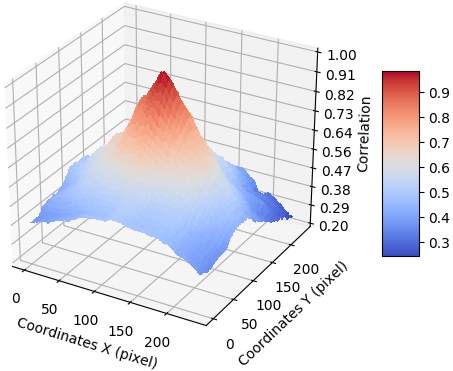}}
   \vspace{-0.3cm}
\caption{Correlation between the center pixel and other pixels of depth map. This correlation is calculated based on the center $240\times240$ region of resized $320\times240$ depth maps from NYU depth V2 dataset \cite{Silberman12:ECCV}.}
\label{fig:depth_correlation}
   \vspace{-0.4cm}
\end{figure}

   \vspace{-0.3cm}
\section{Hierarchical hourglass network}
   \vspace{-0.1cm}
%This section describes the proposed hierarchical hourglass network architecture and a proposed L1-norm scale-invariant loss function.

   \vspace{-0.1cm}
\subsection{Network architecture}
   \vspace{-0.1cm}

% To combine the robustness of traditional stereo depth estimation, 

To estimate full-FoV depth robustly, we propose a hierarchical hourglass network which uses both wide color image and traditional tele-FoV stereo depth as input. 
The proposed network comprises three major modules (as shown in Fig. \ref{fig:flowchart_HieHourglass}): single image depth prediction which infers initial depth from wide color image, 
depth propagation which propagates stereo depth from tele-FoV region to surrounding regions (wherein the stereo-matching depth is obtained by the method in \cite{Hernandez16:PCS} + post processing \cite{Min14:TIP}), and depth combination which fuses the initial depth with the propagated depth to generate final output. The initial depth is also used to fill into the surrounding region of tele-FoV stereo depth, so as to make a complete full-FoV input for depth propagation module. All of these three modules employ the hourglass model \cite{Chen16:NIPS} which has an encoder-decoder structure with skip connections.

In order to illustrate the effectiveness of each module, we compare the intermediate outputs and show them in Fig. \ref{fig:expe_wild_stadium} (c, d and e). 
The single image depth module can predict the global structure but lack of details (Fig. \ref{fig:expe_wild_stadium} (c)), especially for uncommon objects which are not covered by training dataset. 
The depth propagation module would refine the stereo depth at tele-FoV region, at the same time propagate it to surrounding regions, but has slight discontinuity artifact at tele-FoV boundary (Fig. \ref{fig:expe_wild_stadium} (d)). The depth combination module will fuse the initial depth with propagated depth to generate better result, and smooth out the aforementioned discontinuity artifact (Fig. \ref{fig:expe_wild_stadium} (e)).

\begin{figure}[t]
    \begin{minipage}[b]{0.32\linewidth}
      %\centering
      \centerline{\includegraphics[width= 1.2in]{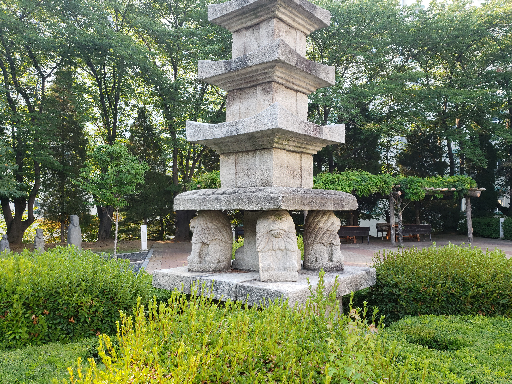}}
      \vspace{0.00cm}
      \centerline{(a)}\medskip
    \end{minipage}
    \hfill
    \begin{minipage}[b]{0.15\linewidth}
     \centerline{\raisebox{0.5\height}{\includegraphics[width= 0.58in]{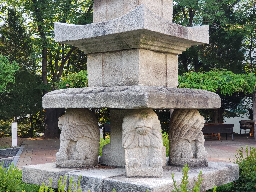}}}
      \vspace{0.00cm}
      \centerline{(b)}\medskip
    \end{minipage}
    \hfill
    \begin{minipage}[b]{0.15\linewidth}
     \centerline{\raisebox{0.5\height}{\includegraphics[width= 0.58in]{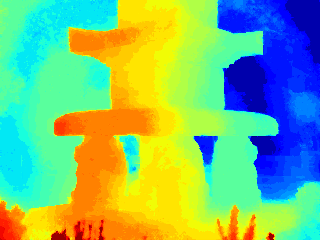}}}
      \vspace{0.00cm}
      \centerline{(c)}\medskip
    \end{minipage}
    \hfill
    \begin{minipage}[b]{0.32\linewidth}
     \centering
      \centerline{\includegraphics[width= 1.2in]{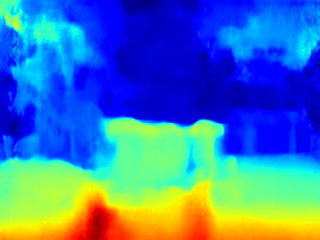}}
      \vspace{0.00cm}
      \centerline{(d)}\medskip
    \end{minipage}

    \begin{minipage}[b]{0.32\linewidth}
      %\centering
      \centerline{\includegraphics[width= 1.2in]{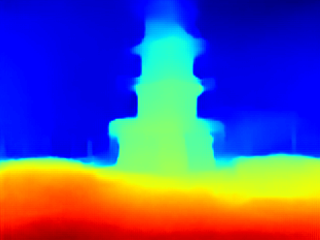}}
      \vspace{0.00cm}
      \centerline{(e)}\medskip
    \end{minipage}
    \hfill
    \begin{minipage}[b]{0.32\linewidth}
      %\centering
      \centerline{\includegraphics[width= 1.2in]{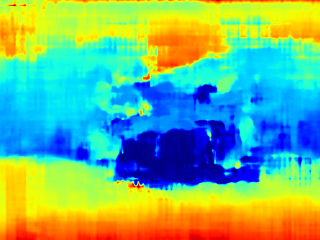}}
      \vspace{0.00cm}
      \centerline{(f)}\medskip
    \end{minipage}
    \hfill
    %\hspace{1.46cm}
    \begin{minipage}[b]{0.32\linewidth}
      %\centering
      \centerline{\includegraphics[width= 1.2in]{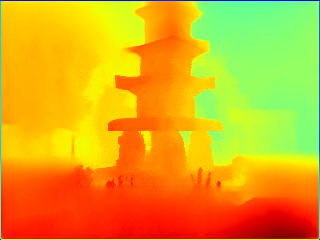}}
      \vspace{0.00cm}
      \centerline{(g)}\medskip
    \end{minipage}

   \vspace{-0.4cm}

    %\centerline{\epsfig{figure=msePS.eps,width=3.0in}}
    \caption {Depth comparisons. (a) Wide image; (b) Tele image; (c) Traditional tele-FoV stereo-matching method: matching cost calculation/optimization \cite{Hernandez16:PCS} + post processing \cite{Min14:TIP}; (d) DNN-based single image depth method in \cite{Godard17:CVPR}; (e) DNN-based single image depth method in \cite{Li18:CVPR}; (f) Pure DNN-based tele-wide stereo matching method in \cite{El-Khamy19:CVPRW}; (g) Our result.} \label{fig:expe_wild_tower}
   \vspace{-0.4cm}
    \end{figure}

   \vspace{-0.3cm}
\subsection{Loss function}
   \vspace{-0.1cm}

To train the whole network efficiently, we construct loss function for each module (as shown in Fig. \ref{fig:flowchart_HieHourglass}), and the final loss is weighted sum of these modules loss functions.

    \begin{equation}
   \vspace{-0.1cm}
    \label{eq:totalLossFunction}
    L = w_{1}L_{1} + w_{2}L_{2} + w_{3}L_{3}
   %\vspace{-0.1cm}
    \end{equation}
where $L$ is the final loss, $L_{1}, L_{2}$ and $L_{3}$ are the loss function of single image depth, depth propagation and depth combination modules, respectively.
$w_{1}, w_{2}$ and $w_{3}$ are the weights, they are set as $0.5$, $0.5$ and {1}, respectively.

For loss function $L_{k}, k=1,2,3$, we propose a L1-norm scale-invariant loss function, 
which regulates predicted log depth to have similar between-points relationships with ground truth.
Compared with widely-used $L_{2}$ norm, $L_{1}$ norm is robust and less sensitive to outliers \cite{Gorard:BJES05}.
It is written as:
    \begin{equation}
   \vspace{-0.1cm}
    \label{eq:L1normScaleInvariant}
    \begin{array}{rl}
    L_{k} &= \frac{1}{N^2} \sum_{i=1}^{N} \sum_{j=1}^{N} \left | \left ( P_{k}^{i} - P_{k}^{j}  \right ) - \left ( T^{i} - T^{j} \right ) \right | \\
            &= \frac{1}{N^2} \sum_{i=1}^{N} \sum_{j=1}^{N} \left | \left ( P_{k}^{i} - T^{i}  \right ) - \left ( P_{k}^{j} - T^{j} \right ) \right | \\
            &= \frac{1}{N^2} \sum_{i=1}^{N} \sum_{j=1}^{N} \left | \left ( D_{k}^{i} - D_{k}^{j}  \right ) \right |
    \end{array}
   %\vspace{-0.1cm}
    \end{equation}
where $P_{k}^{i}$ and $P_{k}^{j}$ are predicted log depth of module $k$ at pixel position $i$ and $j$, respectively. 
$T^{i}$ and $T^{j}$ are ground-truth log depth at pixel position $i$ and $j$, respectively. 
$N$ is the total number of pixels. $D_k$ is the deviation between prediction $P_k$ and ground truth $T$, e.g. $D_{k}=P_{k}-T$. 
Direct calculating the absolute difference of deviations $\left | D_{k}^{i} - D_{k}^{j} \right |$ on all possible pixels pairs is quite time consuming.
To accelerate calculation, we compute absolute difference of deviations only between each pixel and its neighboring pixels, 
because of low correlations between a pixel and other spatially distant pixels of depth map (as shown in Fig. \ref{fig:depth_correlation}).
Then the L1-norm scale-invariant loss function can be rewritten as
    \begin{equation}
   \vspace{-0.2cm}
    \label{eq:L1normScaleInvariant_appro}
    L_{k} = \frac{1}{N \times M} \sum_{i=1}^{N} \sum_{m=1}^{M}  \left |  D_{k}^{i} - D_{k}^{im}  \right |
%\sum_{j=1}^{M} 
   %\vspace{-0.1cm}
    \end{equation}
where $m$ is the neighboring pixels index of the pixel $i$, and $M$ is the number of neighboring pixels. 
Absolutely the larger neighborhood would produce better results. 
Considering time complexity, we set neighborhood as a $17\times17$ window.

    \begin{table}[t]
    \caption{Performance comparison on KITTI dataset. RMSE: root mean squared error; REL: mean absolute relative error; $\delta_{i}$: percentage of predicted pixels where the relative error is within a threshold $1.25^{i}$ \cite{Ma18:ICRA}.}
    \centering
    \scalebox{1}{
    \begin{tabular}{|c|c|c|c|c|c|}
    \hline
     &\multicolumn{2}{|c|}{Lower is better}&\multicolumn{3}{|c|}{Higher is better}\\
    \hline
     Method&RMSE&REL&$\delta_{1}$&$\delta_{2}$&$\delta_{3}$\\
    \hline
    Mancini \cite{Mancini16:IROS}&7.508& - &31.8&61.7&81.3 \\
    \hline
    Eigen et al. \cite{Eigen14:NIPS}&7.156&0.190&69.2&89.9&96.7\\
    \hline
    Ma et al. \cite{Ma18:ICRA}&6.266&0.208&59.1&90.0&96.2\\
    \hline
    Godard et al. \cite{Godard17:CVPR}&5.927&0.148&80.3&92.2&96.4\\
    \hline
    Our method&\textbf{2.440}&\textbf{0.05}&\textbf{95.2}&\textbf{98.3}&\textbf{99.3}\\
    \hline
    \end{tabular}
    }
    \label{tab:KITTI_objectiveComp}
   \vspace{-0.4cm}
    \end{table}

To justify the effectiveness of the proposed L1-norm scale-invariant loss function, we compare it with the result of our network which employs widely-used L2-norm scale-invariant loss function \cite{Eigen14:NIPS}, as shown in Fig. \ref{fig:expe_wild_stadium} (e, f). It can be observed that the L1-norm loss function can generate better global structure, especially for background.

   \vspace{-0.3cm}
\section{Experiments}
   \vspace{-0.2cm}

In order to validate the effectiveness and robustness of our proposed network, firstly we evaluate our method on several wild test images, which covers various different scenarios. The network is trained on MegaDepth dataset \cite{Li18:CVPR}, where the tele-FoV depth for training is cropped from ground truth. Then we evaluate our network on two standard depth datasets: KITTI \cite{Geiger13:IJRR} and NYU Depth V2 \cite{Silberman12:ECCV}, where we train our network on their training set, respectively.

   \vspace{-0.4cm}
\subsection{Wild test images}
   \vspace{-0.2cm}

We capture several test images at various scenarios by tele-wide camera of Galaxy S9 plus. MegaDepth \cite{Li18:CVPR} is employed as training dataset, which contains 200 landmarks around the world, and there are totally 100K images which have Euclidean depth data. 
All of the images with their depth are scaled to $240\times320$, and the center $120\times160$ region is set as tele FoV. 
The training epoch number is set as 20. During test stage, the tele-FoV stereo-matching depth is obtained by the traditional stereo matching method \cite{Hernandez16:PCS} + post processing \cite{Min14:TIP}.

%And there are totally 130K photos, where 100K images have Euclidean depth data, and 30K images have ordinal foreground/background information.

%Because all the ground-truth depth of this dataset are reconstructed based on structure-from-motion [25] and mutli-view-stereo methods [26] from internet photos, their magnitude may have various %scale factor. 

The proposed network is compared with state-of-the-art single image depth prediction methods \cite{Godard17:CVPR, Li18:CVPR} and a pure DNN-based tele-wide stereo matching method \cite{El-Khamy19:CVPRW}. All of the test codes of these methods are obtained either from their official websites or from authors. The comparison results are shown in Fig. \ref{fig:expe_wild_statue} and  Fig. \ref{fig:expe_wild_tower}. It can be observed that our method can achieve better and more robust subjective quality, especially for the foreground objects.

% Besides Fig. \ref{fig:expe_wild_statue}, some of comparisons are shown in Fig. \ref{fig:expe_wild_bottle} and Fig. \ref{fig:expe_wild_man}. With the support of traditional stereo tele-FoV depth, 

   \vspace{-0.4cm}
\subsection{Standard dataset}
   \vspace{-0.15cm}

KITTI dataset \cite{Geiger13:IJRR} contains outdoor scenes with resolution $375\times1241$ captured by cameras and depth sensors on a driving car. We use the 22600 training images from 28 scenes and 697 test images from another 29 scenes based on Eigen split \cite{Eigen14:NIPS}. A $256\times1216$ region is horizontally-random and vertically-bottom cropped from each image for training and testing, wherein the center $128\times608$ region of ground truth is used as tele-FoV depth for training. Our model is trained with 40 epochs. To clearly know the maximum benefit of our method regardless of stereo depth quality, we use the center $128\times608$ region of ground-truth depth as tele-FoV depth for test. Because all of the training and testing images are captured by the same device, we can combine the proposed L1-norm scale-invariant loss function in Equ. \ref{eq:L1normScaleInvariant_appro} with the common L1 norm loss function to get better results: $L_{k}=0.5 \times \frac{1}{N \times M} \sum_{i=1}^{N} \sum_{m=1}^{M}  \left |  D_{k}^{i} - D_{k}^{im}  \right | + \frac{1}{N} \sum_{i=1}^{N} \left |  D_{k}^{i} \right |$, wherein $D$ is the deviation between predicted depth and ground-truth depth.

\begin{figure}[t]
    \begin{minipage}[b]{0.98\linewidth}
      %\centering
      \centerline{\includegraphics[width= 2.7in]{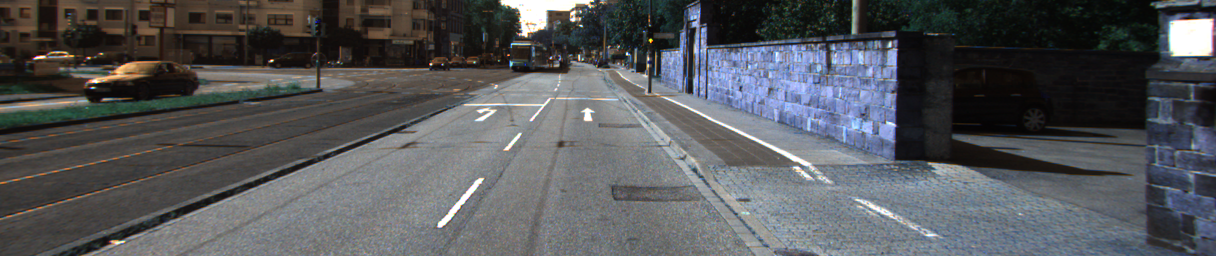}}
      \vspace{0.00cm}
      %\centerline{(a)}\medskip
    \end{minipage}
    %\hfill
    \begin{minipage}[b]{0.98\linewidth}
      %\centering
      \centerline{\includegraphics[width= 2.7in]{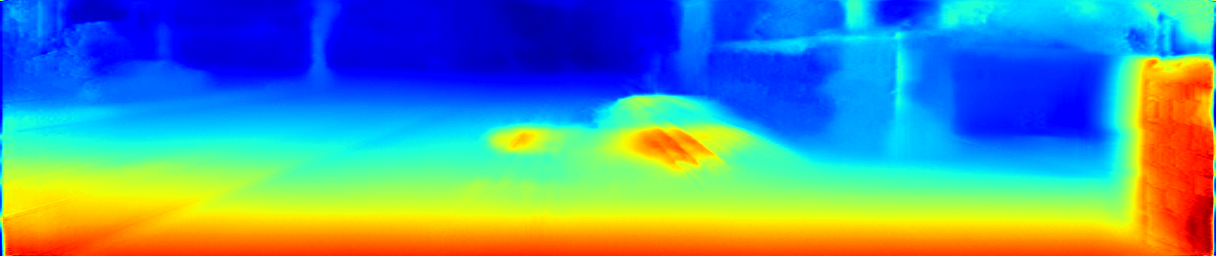}}
      \vspace{0.00cm}
      %\centerline{(b)}\medskip
    \end{minipage}

    \begin{minipage}[b]{0.98\linewidth}
      %\centering
      \centerline{\includegraphics[width= 2.7in]{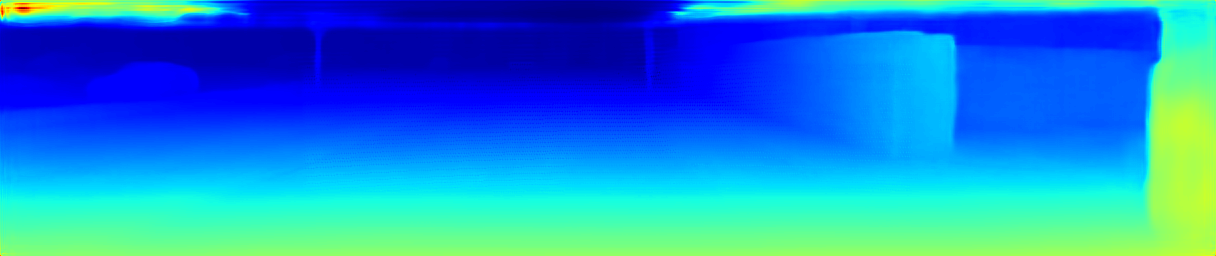}}
      \vspace{0.00cm}
      %\centerline{(c)}\medskip
    \end{minipage}
    %\hfill
    %\hspace{1.46cm}
    \begin{minipage}[b]{0.98\linewidth}
      %\centering
      \centerline{\includegraphics[width= 2.7in]{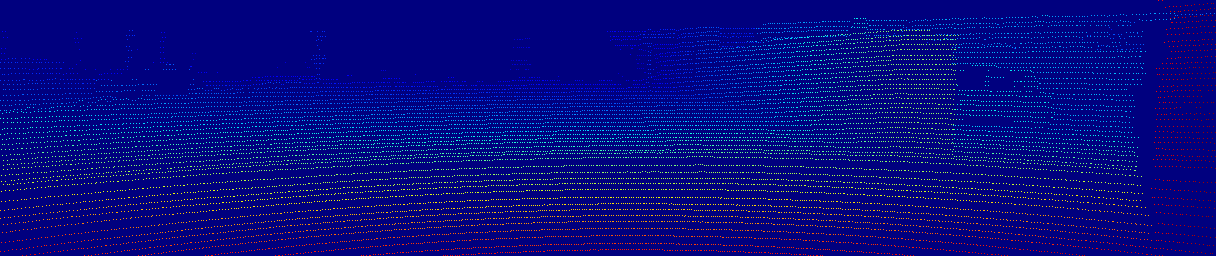}}
      \vspace{0.00cm}
      %\centerline{(d)}\medskip
    \end{minipage}

   \vspace{-0.4cm}

    \caption {Depth comparisons on KITTI test images. From top to bottom: Reference wide image, DNN method in \cite{Godard17:CVPR}, Our result, Ground truth.} \label{fig:kitti_subjective}
   \vspace{-0.4cm}
    \end{figure}

The objective and subjective comparisons with state-of-the-art single image depth prediction methods on KITTI test set are shown in Tab. \ref{tab:KITTI_objectiveComp} and Fig. \ref{fig:kitti_subjective}, respectively. Even though the training and test data are not the same across various methods, the scenes are similar because they are captured during driving and from the same sensor \cite{Ma18:ICRA}. We can see that our model with the support of tele-FoV depth can boost RMSE, REL and $\delta$ values a lot, and also achieve better subjective quality.

  \begin{table}[t]
    \caption{Performance comparison on NYU Depth V2 dataset.}
    \centering
    \scalebox{1}{
    \begin{tabular}{|c|c|c|c|c|c|}
    \hline
     &\multicolumn{2}{|c|}{Lower is better}&\multicolumn{3}{|c|}{Higher is better}\\
    \hline
     Method&RMSE&REL&$\delta_{1}$&$\delta_{2}$&$\delta_{3}$\\
    \hline
    Roy et al. \cite{Roy16:CVPR}&0.744&0.187&-&-&- \\
    \hline
    Eigen et al. \cite{Eigen14:NIPS}&0.641&0.158&76.9&95.0&98.8\\
    \hline
    Laina et al. \cite{Laina16:3DV}&0.573&0.127&81.1&95.3&98.8\\
    \hline
    Ma et al. \cite{Ma18:ICRA}&0.514&0.143&81.0&95.9&98.9\\
    \hline
    Our method&\textbf{0.334}&\textbf{0.087}&\textbf{92.4}&\textbf{98.0}&\textbf{99.4}\\
    \hline
    \end{tabular}
    }
    \label{tab:NYU_objectiveComp}
   \vspace{-0.3cm}
    \end{table}

\begin{figure}[t]
    \begin{minipage}[b]{0.49\linewidth}
      %\centering
      \centerline{\includegraphics[width= 1.6in]{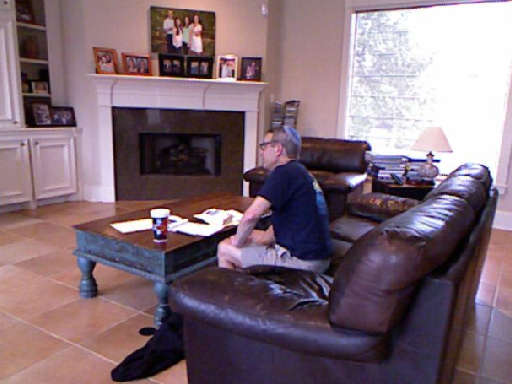}}
      \vspace{0.00cm}
      \centerline{(a)}\medskip
    \end{minipage}
    \hfill
\hspace{-0.3cm}
    \begin{minipage}[b]{0.49\linewidth}
      %\centering
      \centerline{\includegraphics[width= 1.6in]{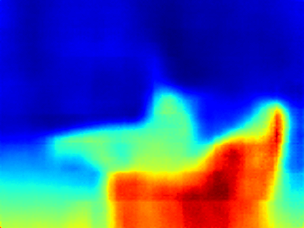}}
      \vspace{0.00cm}
      \centerline{(b)}\medskip
    \end{minipage}
  \vspace{-0.3cm}
    \begin{minipage}[b]{0.49\linewidth}
      %\centering
      \centerline{\includegraphics[width= 1.6in]{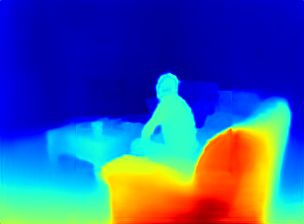}}
      \vspace{0.00cm}
      \centerline{(c)}\medskip
    \end{minipage}
    \hfill
\hspace{-0.3cm}
    %\hspace{1.46cm}
    \begin{minipage}[b]{0.49\linewidth}
      %\centering
      \centerline{\includegraphics[width= 1.6in]{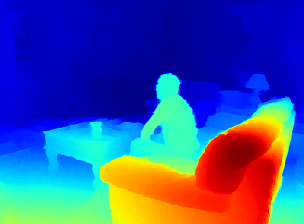}}
      \vspace{0.00cm}
      \centerline{(d)}\medskip
    \end{minipage}

   \vspace{-0.3cm}

    \caption {Depth comparisons on NYU test images. (a) Reference image; (b) DNN method in \cite{Ma18:ICRA}: RGB image as input; (c) Our result: RGB image + tele-FoV depth as input; (d) Ground truth.} \label{fig:nyu_subjective}
   \vspace{-0.4cm}
    \end{figure}

% \subsection{NYU dataset}

The NYU Depth V2 dataset \cite{Silberman12:ECCV} consists of color images and their depth maps captured by Microsoft Kinect from 464 various indoor scenes. 
Based on official split, we use 249 scenes for training, and 654 images \cite{Eigen14:NIPS, Laina16:3DV} of the rest 215 scenes for testing. 
All of the images with their depth maps are scaled to $224\times304$, and the center $112\times152$ region of depth is used as tele-FoV depth for training. 
We use ground-truth tele-Fov depth together with color image as input for test, not only because the NYU dataset has only single image without stereo pairs, 
but also because we want to know the maximum benefit of our model regardless of stereo depth quality. We trained our model with 40 epochs. 
Because all of the training and testing images of NYU dataset are captured by the same device, we can use the combined loss function from the proposed L1-norm scale-invariant loss and the common L1 norm loss (same as KITTI dataset). 
The objective and subjective comparisons with state-of-the-art single image depth prediction methods are shown in Tab. \ref{tab:NYU_objectiveComp} and Fig. \ref{fig:nyu_subjective}, respectively. The comparisons demonstrate that our model can obtain better RMSE, REL, $\delta$ values as well as better subjective quality.

    \vspace{-0.5cm}
\section{Conclusion}
   \vspace{-0.3cm}

We introduced a hierarchical hourglass network for robust full-FoV depth estimation in tele-wide camera system, which combines the robustness of traditional stereo-matching methods with the accuracy of DNN methods. Experiments demonstrate its robustness and better quality in both subjective and objective evaluations. We believe this new method opens up a door for research on combining robustness of traditional signal processing into deep learning for depth estimation. In the future, we will investigate new network structure and extend our framework into other computer vision problems.

%\section*{References}
%\section{REFERENCES}
%\label{sec:refs}

\bibliographystyle{IEEEbib}
\bibliography{refs_1}

\end{document}